# Lookahead optimizer improves the performance of Convolutional Autoencoders for reconstruction of natural images

Sayan Nag

nagsayan112358@gmail.com

**Abstract**

Autoencoders are a class of artificial neural networks which have gained a lot of attention in the recent past. Using the encoder block of an autoencoder the input image can be compressed into a meaningful representation. Then a decoder is employed to reconstruct the compressed representation back to a version which looks like the input image. It has plenty of applications in the field of data compression and denoising. Another version of Autoencoders (AE) exist, called Variational AE (VAE) which acts as a generative model like GAN. Recently, an optimizer was introduced which is known as lookahead optimizer which significantly enhances the performances of Adam as well as SGD. In this paper, we implement Convolutional Autoencoders (CAE) and Convolutional Variational Autoencoders (CVAE) with lookahead optimizer (with Adam) and compare them with the Adam (only) optimizer counterparts. For this purpose, we have used a movie dataset comprising of natural images for the former case and CIFAR100 for the latter case. We show that lookahead optimizer (with Adam) improves the performance of CAEs for reconstruction of natural images.

## 1. Introduction

In the past few years, Deep Learning has made significant progress in the field of computer vision [1-12]. One of the important architectures of unsupervised learning is autoencoder [8] which has drawn of lot of attention in the field of computer vision. The job of an autoencoder is to transform the input to an output with least distortion which helps in data compression, removal of redundant variables and data denoising [14-15]. Convolutional Neural Networks have shown significant improvement in object recognition and classification tasks. The power of convolution has been used to leverage the performance of vanilla Autoencoder eventually giving rise to Convolutional Autoencoder (CAE) [16-17].

On the other hand, significant developments have been made in the field of optimization [18-19]. One such recent advancement is lookahead optimizer which improves the performance of SGD and Adam and their respective variants [19]. Merging these two, we demonstrate that CAEs perform better with lookahead optimizer compared to the Adam counterpart for a movie dataset comprising of natural images. We also show that CVAEs [20] perform better with lookahead optimizer as well for CIFAR100 images [13].

Our paper has been organized as follows: Section 2 represents the methods used in the study, Section 3 demonstrates the experiments and the results section and finally Section 3 contains the conclusion.

## 2. Methods

### 2.1. Dataset

The datasets used in the paper are: (i) CIFAR100 [13], (ii) movie dataset consisting of 108000 training RGB images of size 128 x 128 x 3 and 8100 validation images of the same size [21-22]. Fig. 1 shows the sample images from both training and validation datasets. For this work, we have resized each of the images to 64 x 64x 3.

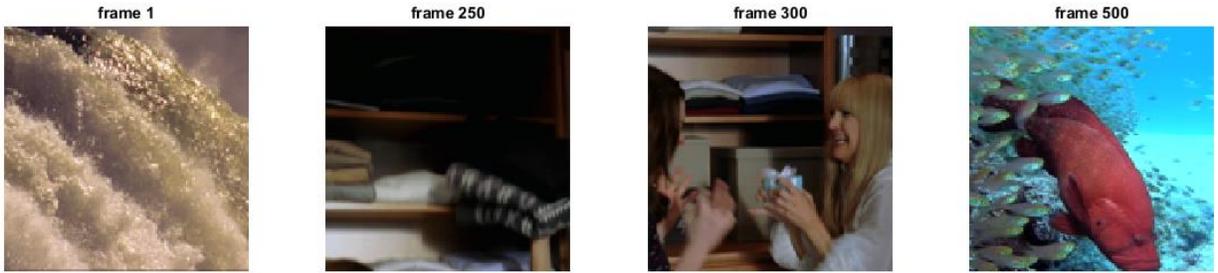

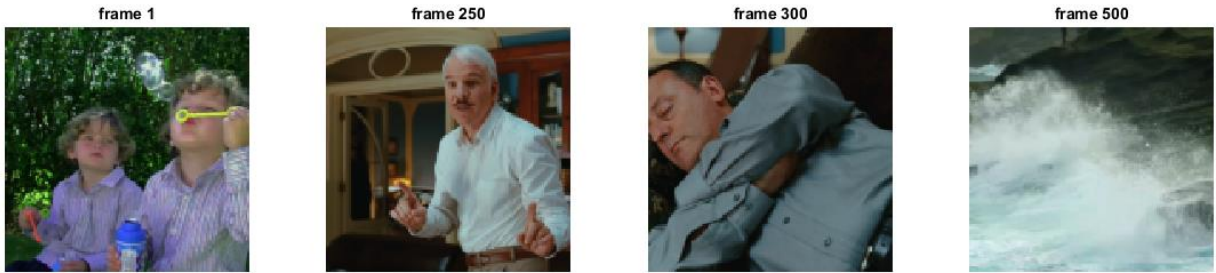

Fig 1. (a) sample images from the training dataset and (b) sample images from the validation dataset

## 2.2. Architecture

**Autoencoder (AE):** A vanilla autoencoder is a three-layered neural network where the connections are fully connected. It consists of an encoder, a hidden layer or node or bottleneck (compressed representation of input) and a decoder.

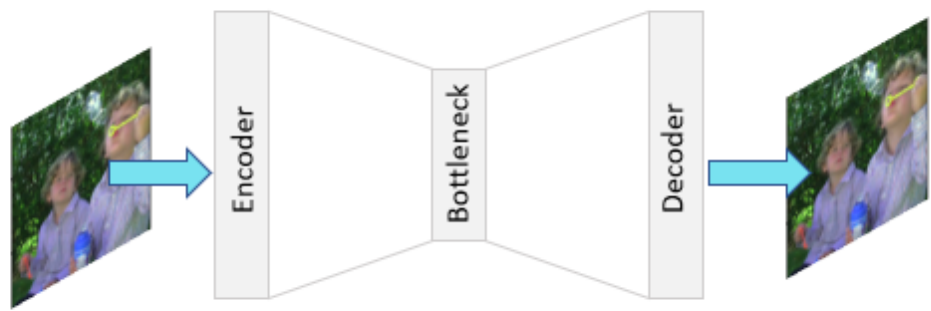

Fig 2. Example Autoencoder with Encoder, Bottleneck, Decoder

**Convolutional Autoencoders (CAE):** An enhancement of the vanilla autoencoder is Convolutional autoencoder where the fully connected layers are modified to convolutional layers. The encoder consists of convolutional layers and the decoder comprises of transposed convolutional layers.

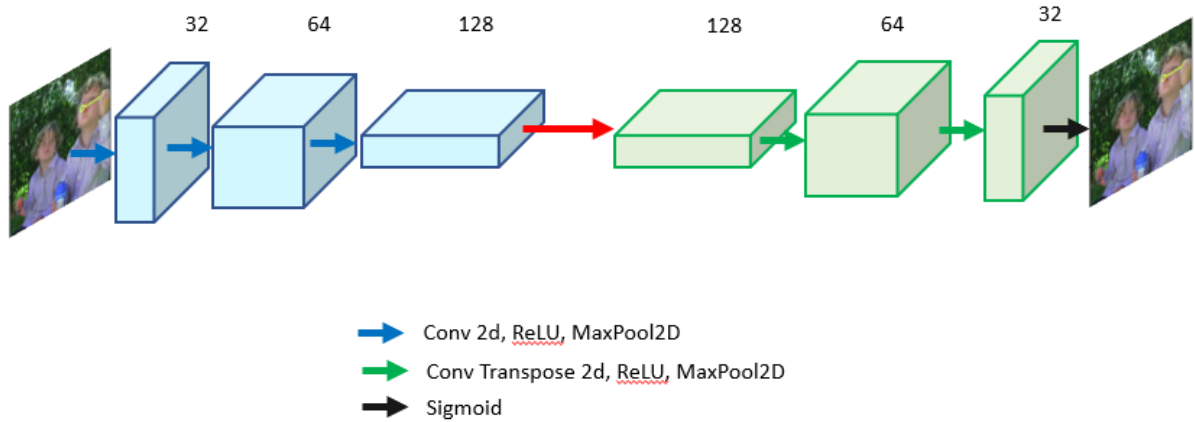

Fig 3. Convolutional Autoencoder

**Convolutional Variational Autoencoder (CVAE):** CVAE extends the basic structure of VAE. It comprises of an encoder consisting of convolutional layers, a bottleneck layer composed of mean and variance layers which together give rise of latent space representation layer, and finally a decoder which is composed of transposed convolutional layers.

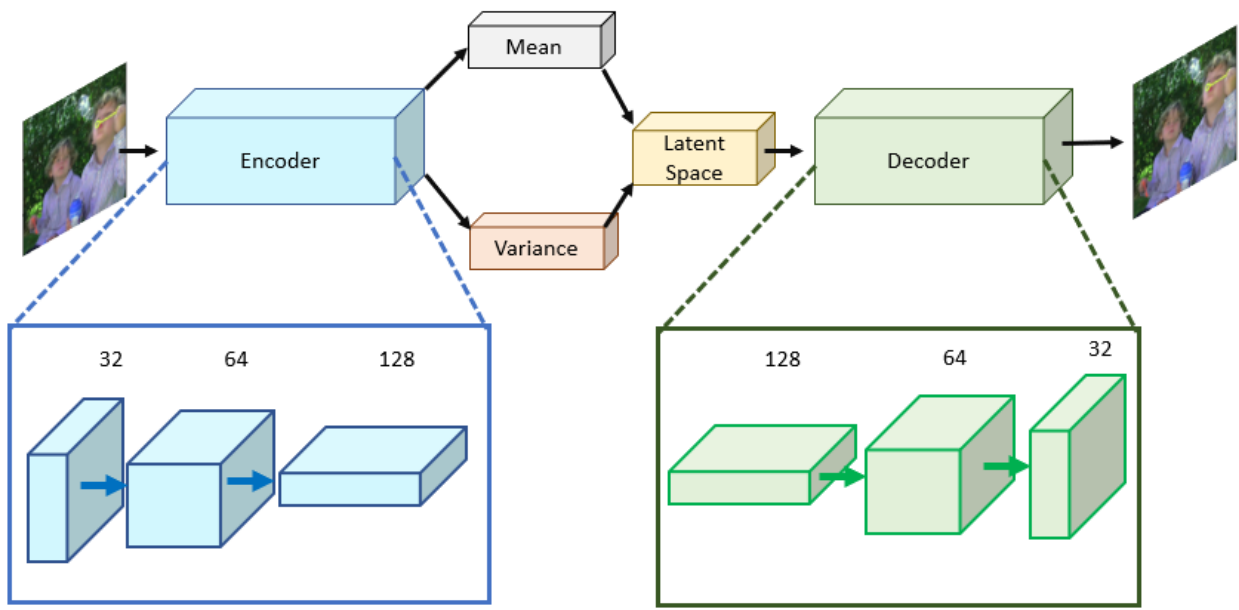

Fig 4. Convolutional Variational Autoencoder

## 3. Experiments and Results

For this work, each of the movie image frames have been resized to 64 x 64 x 3 from 128 x 128 x 3. We have considered CIFAR100 for training our CVAE and the movie data for training our CAE. We have used two optimizers in our experiment, namely Adam and Lookahead with Adam and we demonstrate that lookahead optimizer improves the performance of both CAE and CVAE. We have considered MSE loss for CAE and for CVAE we have considered a combination of Binary Cross Entropy and KL Divergence losses [23-24]. Figs 5-8 show the actual and reconstructed images obtained from the respective architectures using Adam and Lookahead with Adam optimizers. Fig 9 shows the training loss curves over 50 epochs. It can be seen that autoencoders with lookahead optimizer (with Adam) perform better compared to Adam (only) counterparts.

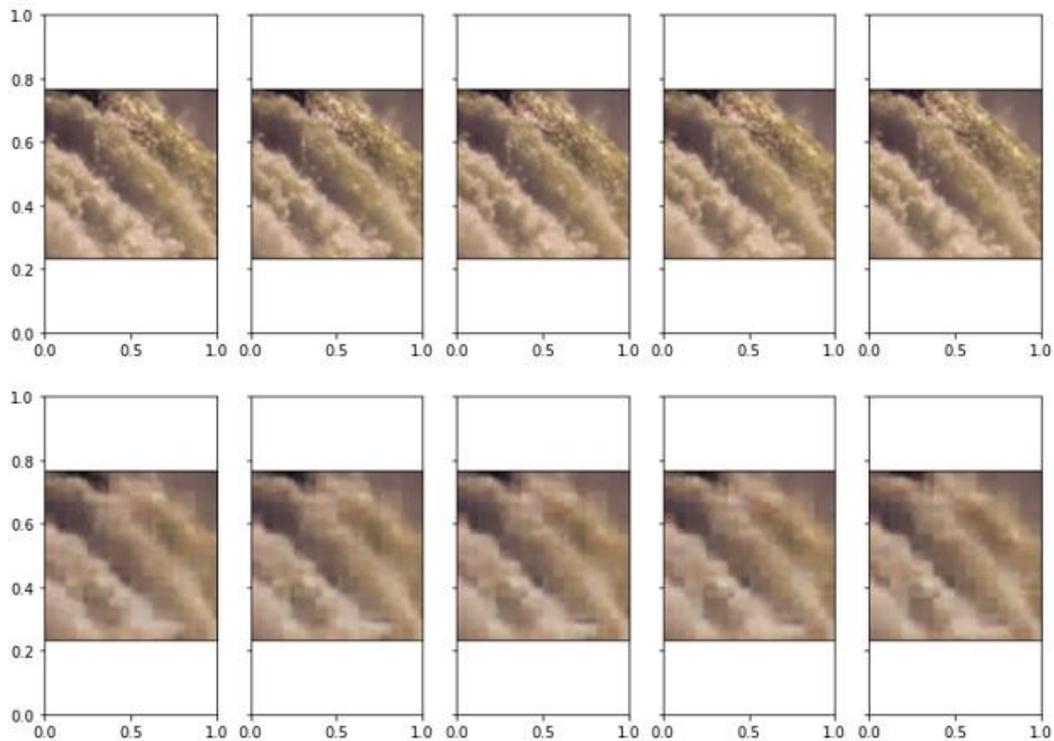

Fig 5. Actual (top) and Reconstructed (bottom) images of first five frames of the movie data using Adam as an optimizer for CAE

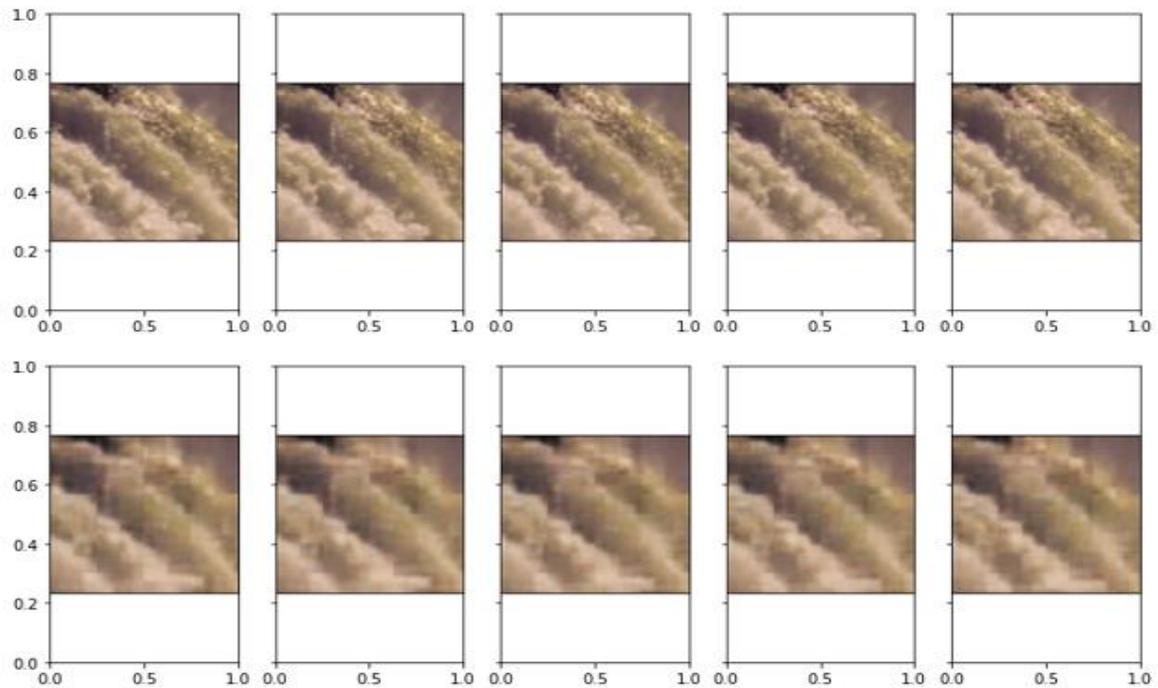

Fig 6. Actual (top) and Reconstructed (bottom) images of first five frames of the movie data using Lookahead with Adam as an optimizer for CAE

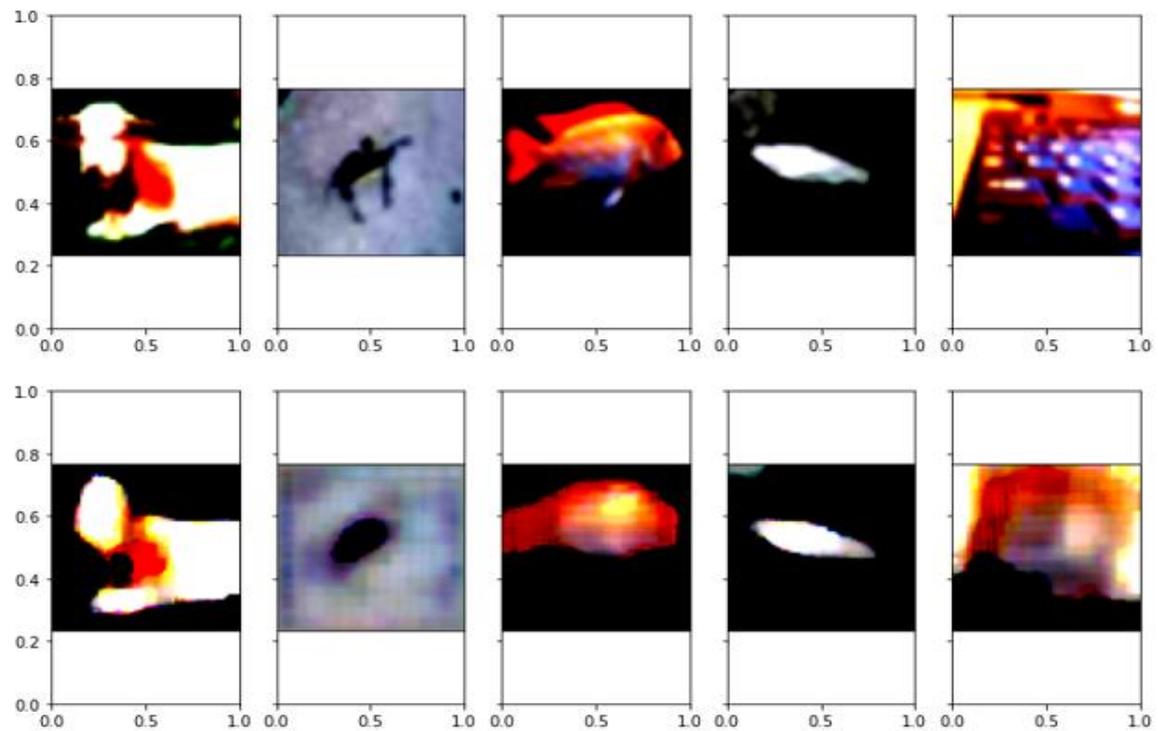

Fig 7. Actual (top) and Reconstructed (bottom) images of five sample CIFAR-100 data using Adam as an optimizer for CVAE

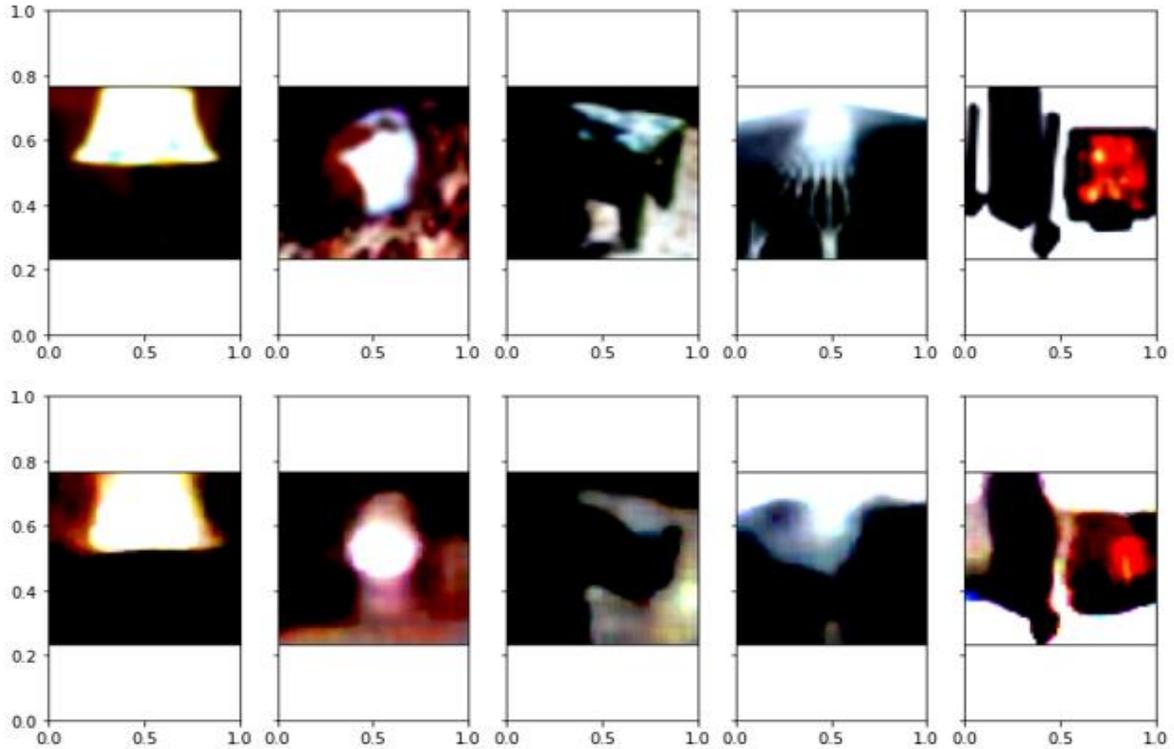

Fig 8. Actual (top) and Reconstructed (bottom) images of five sample CIFAR-100 data using Lookahead with Adam as an optimizer for CVAE

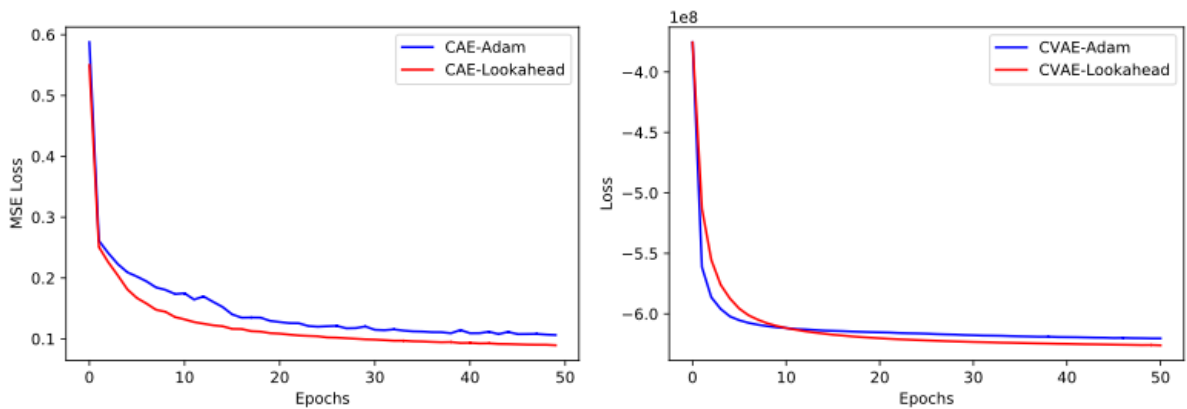

Fig 9. Training loss curves for CAE (left) and CVAE (right)

## 4. Conclusion

The results shown above clearly indicate the advantage of using lookahead optimizer with Adam for training convolutional autoencoders with images from the movie dataset and as well as CIFAR-100 dataset. The results are not state-of-the-art because the main goal of this study is to only demonstrate the significance of lookahead optimizer, and the state-of-the-art architectures were not explored. The loss curves for validation and testing losses are missing because we just wanted to demonstrate the impact of lookahead optimizer on training loss. In future, this work can be explored with skip-connected encoder-decoder architectures.

# References

1. Krizhevsky, A., Sutskever, I., & Hinton, G. E. Imagenet classification with deep convolutional neural networks. In Advances in neural information processing systems (pp. 1097-1105)(2012).
2. LeCun, Y. (2015). LeNet-5, convolutional neural networks. URL: http://yann. lecun. com/exdb/lenet, 20.
3. LeCun, Yann, Yoshua Bengio, and Geoffrey Hinton. "Deep learning." *nature* 521.7553 (2015): 436-444.
4. Goodfellow, Ian, et al. "Generative adversarial nets." *Advances in neural information processing systems*. 2014.
5. Goodfellow, Ian, et al. *Deep learning*. Vol. 1. No. 2. Cambridge: MIT press, 2016.
6. Xu, Kelvin, et al. "Show, attend and tell: Neural image caption generation with visual attention." *International conference on machine learning*. 2015.
7. Kingma, Diederik P., and Max Welling. "Auto-encoding variational bayes." *arXiv preprint arXiv:1312.6114* (2013).
8. Baldi, P. Autoencoders, unsupervised learning, and deep architectures. In Proceedings of ICML Workshop on Unsupervised and Transfer Learning (pp. 37-49)(2012, June).
9. Chen, Ricky TQ, et al. "Neural ordinary differential equations." Advances in neural information processing systems. 2018.
10. Hammernik, Kerstin, et al. "Learning a variational network for reconstruction of accelerated MRI data." *Magnetic resonance in medicine* 79.6 (2018): 3055-3071.
11. Nitski, Osvald, et al. "CDF-Net: Cross-Domain Fusion Network for Accelerated MRI Reconstruction." *International Conference on Medical Image Computing and Computer-Assisted Intervention*. Springer, Cham, 2020.
12. Bhattacharyya, Mayukh, and Sayan Nag. "Hybrid Style Siamese Network: Incorporating style loss in complimentary apparels retrieval." *arXiv preprint arXiv:1912.05014* (2019).
13. Alex Krizhevsky and Geoffrey Hinton. Learning multiple layers of features from tiny images. Technical report, University of Toronto, 2009
14. Vincent, Pascal, et al. "Stacked denoising autoencoders: Learning useful representations in a deep network with a local denoising criterion." *Journal of machine learning research* 11.12 (2010).
15. Vincent, Pascal, et al. "Extracting and composing robust features with denoising autoencoders." *Proceedings of the 25th international conference on Machine learning*. 2008.
16. Masci, Jonathan, et al. "Stacked convolutional auto-encoders for hierarchical feature extraction." *International conference on artificial neural networks*. Springer, Berlin, Heidelberg, 2011.
17. Zhang, Yifei. "A better autoencoder for image: Convolutional autoencoder." *ICONIP17-DCEC. Available online: http://users. cecs. anu. edu. au/Tom. Gedeon/conf/ABCs2018/paper/ABCs2018_paper_58. pdf (accessed on 23 March 2017)*. 2018.
18. Kingma, Diederik P., and Jimmy Ba. "Adam: A method for stochastic optimization." *arXiv preprint arXiv:1412.6980* (2014).
19. Zhang, Michael, et al. "Lookahead optimizer: k steps forward, 1 step back." *Advances in Neural Information Processing Systems*. 2019.
20. Pu, Yunchen, et al. "Variational autoencoder for deep learning of images, labels and captions." *Advances in neural information processing systems*. 2016.
21. Nishimoto, S., Vu, A. T., Naselaris, T., Benjamini, Y., Yu, B., & Gallant, J. L. (2011). Reconstructing visual experiences from brain activity evoked by natural movies. Current Biology, 21(19), 1641-1646.
22. Shinji Nishimoto, An T. Vu, Thomas Naselaris, Yuval Benjamini, Bin Yu, Jack L. Gallant (2014): Gallant Lab Natural Movie 4T fMRI Data. CRCNS.org. http://dx.doi.org/10.6080/K00Z715X
23. Kullback, S.; Leibler, R.A. (1951). "On information and sufficiency". Annals of Mathematical Statistics. 22 (1): 79–86
24. MacKay, David JC, and David JC Mac Kay. *Information theory, inference and learning algorithms*. Cambridge university press, 2003.